\documentclass[final]{cvpr}

\usepackage{times}
\usepackage{epsfig}
\usepackage{graphicx}
\usepackage{amsmath}
\usepackage{amssymb}
\usepackage{algorithm}  
\usepackage{algorithmic}
\usepackage{multirow}
\usepackage{booktabs}
\usepackage[symbol,hang,flushmargin]{footmisc}

\usepackage{paralist}

\usepackage{cuted}

\usepackage[pagebackref=true,breaklinks=true,colorlinks,bookmarks=false]{hyperref}

\def\cvprPaperID{5064} 
\def\confYear{CVPR 2021}

\newcommand{\mname}{MTUNet}
\newcommand{\Mname}{MTUNet}

\newcommand{\cname}{PM}

\begin{document}

\title{Match Them Up: Visually Explainable Few-shot Image Classification}

\author{Bowen Wang$^{1,3}$, Liangzhi Li$^{1,5}$, Manisha Verma$^{1,5}$, Yuta Nakashima$^{1,5}$,\\Ryo Kawasaki$^{2,4}$, Hajime Nagahara$^{1,5}$\\
$^1$Institute for Datability Science (IDS) $\quad$ $^2$Graduate School of Medicine\\
Osaka University, Japan\\
{\tt\small $^3$bowen.wang@is.ids.osaka-u.ac.jp} {\tt\small $^4$ryo.kawasaki@ophthal.med.osaka-u.ac.jp}\\
{\tt\small $^5$\{li, mverma, n-yuta, nagahara\}@ids.osaka-u.ac.jp}
}

\maketitle
\renewcommand*{\thefootnote}{\arabic{footnote}}
\setcounter{footnote}{5}
\begin{abstract}
Few-shot learning (FSL) approaches are usually based on an assumption that the pre-trained knowledge can be obtained from base (seen) categories and can be well transferred to novel (unseen) categories. However, there is no guarantee, especially for the latter part. This issue leads to the unknown nature of the inference process in most FSL methods, which hampers its application in some risk-sensitive areas. In this paper, we reveal a new way to perform FSL for image classification, using visual representations from the backbone model and weights generated by a newly-emerged explainable classifier. The weighted representations only include a minimum number of distinguishable features and the visualized weights can serve as an informative hint for the FSL process. Finally, a discriminator will compare the representations of each pair of the images in the support set and the query set. Pairs with the highest scores will decide the classification results. Experimental results prove that the proposed method can achieve both good accuracy and satisfactory explainability on three mainstream datasets. 
Code is available
\footnote{https://github.com/wbw520/MTUNet}.
\end{abstract}

\begin{figure}[!t]
	\centering
	\includegraphics[width=\columnwidth]{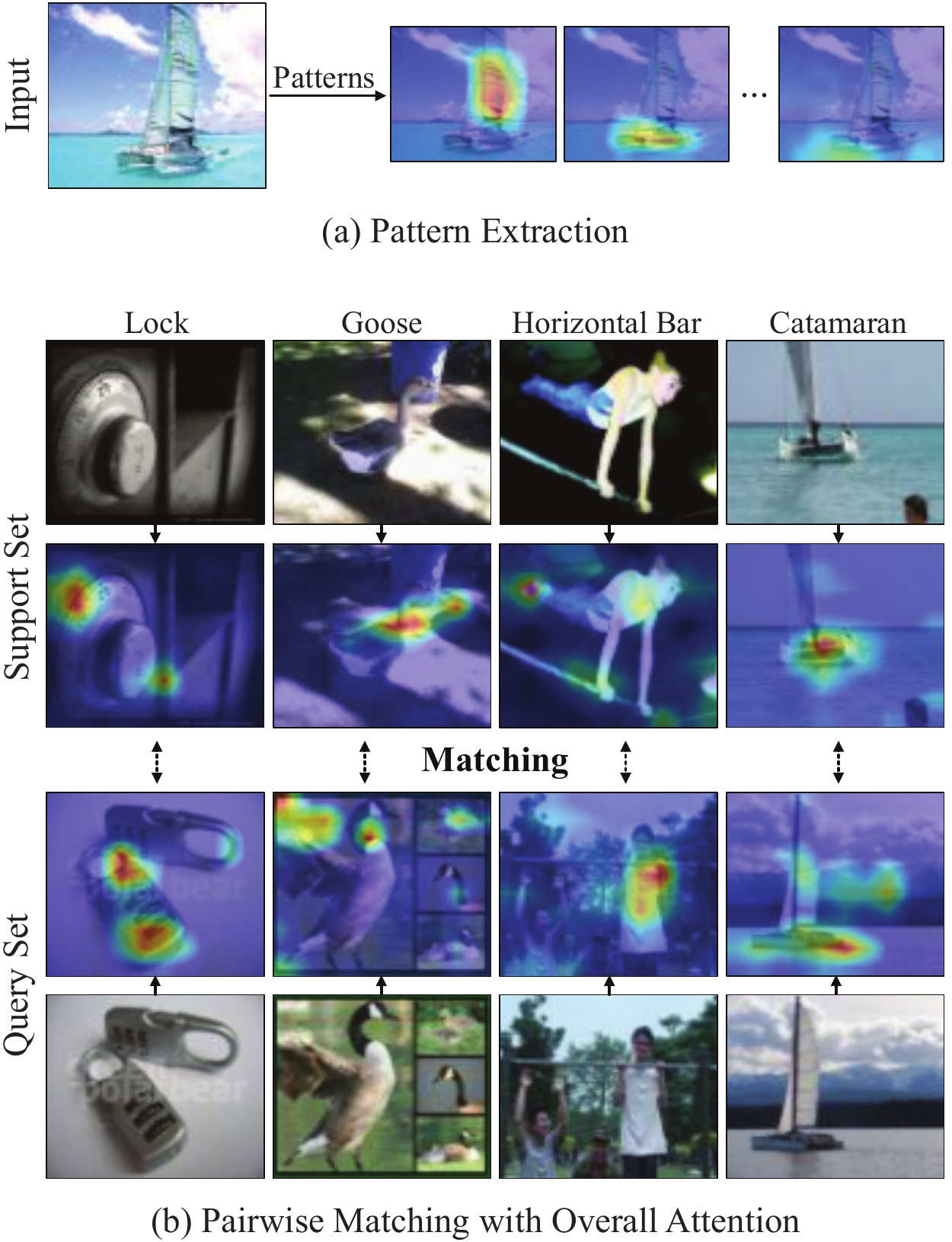}
	\caption{Few-shot learning by pair-matching with the  \textit{pattern extractor} (PE). Images are from mini-ImageNet dataset \cite{vinyals2016matching}.}
	\label{fig_story}
\end{figure}

\section{Introduction}
Few-shot learning (FSL) is of great significance at least for the following two scenarios \cite{FSL_survey}: Firstly, FSL can relieve the heavy needs for data gathering and labeling, which can boost the ubiquitous use of deep learning techniques, especially for users without enough resources. Secondly, FSL is an important solution for applications in which rare cases matter or image acquisition is costly because of high operation difficulty or ethical issues. Typical examples of such applications include computer assisted diagnosis with medical imaging, classification of endangered species, \etc.

There have been lots of FSL methods, most of which are based on the assumption that knowledge can be well extracted from base (seen) classes and transferred to novel (unseen) classes. However, this is not always the case. The knowledge in a pre-trained backbone convolutional neural network (CNN), which computes features of an input image, may sometimes be useless when novel categories have significant visual differences from images of the base categories \cite{yue2020interventional}. 
What makes matter worse is that we even have no way to see if the visual differences between the base and novel categories are significant \textit{for an FSL model}. This raised one essential question: \textit{Is there any way to see what is actually transferred?}


Actually, in the FSL task, most works only treat the convolutional layer as the image embedding tool, and do not pay attention to the reasons for the extracted features. In this paper, we redesign the mechanism of knowledge transfer for FSL tasks, which offers an answer to the above question. Our approach is inspired by what human beings actually do when trying to recognize a rarely seen object. That is, we usually try to find some patterns in the object and match them in a small number of seen examples in our memory. We adopt a recently-emerged explainable classifier, called SCOUTER \cite{li2020scouter}, and propose a new FSL method, named match-them-up network (\mname{}) consisted of \textit{pattern extractor} (PE) and \textit{pairwise matching} (PM).

PE is designed for finding discriminative and consistent patterns for image representation. The knowledge transferred from the base categories to the novel categories is the learned patterns. Owing to the explainability of SCOUTER, the extracted patterns themselves can be easily visualized by exemplifying them in the images as shown in Figure \ref{fig_story}(a). This directly means that we have a way to see what is actually transferred in our FSL pipeline. The patterns extracted in each of the support and query images are aggregated to form discriminative image representation (overall attention), which is used for matching. As shown in Figure \ref{fig_story}(b), the visualization of aggregated patterns collectively shows a consistent and meaningful clue for the images of the same category. For example, PE shows strong attention on the neck of the goose in the second column, which is consistent in both support and query images. Image representation based on the patterns from base categories makes matching between a pair of images much easier by incorporating only a small number of regions to pay attention to.

On top of the PE, PM is adopted to judge whether image pairs belong to the same category or not. Each pair consists of one image from a \textit{support} set and one image from a \textit{query} set. The category of the support image that gives the highest score is regarded as the query image's category. Together with PE, \Mname{} can provide a matching score matrix to further relate the visualization and the model decision.

The main contributions of our work include:
\begin{compactitem}
	\item a new FSL method that can output visual explanations besides classification results to find potential failures of the method,
	\item a new image representation based on filtering original image features, given by a backbone CNN, to keep only informative regions, and relate the visualization with the model decision.
	\item a new light-weighted model to perform accurate FSL image classification.
\end{compactitem}

\section{Related Work}
\subsection{Few-shot Learning}
Recently, deep neural networks have achieved outstanding performance in various classification tasks, thanks to the availability of a sufficient number of images for each category. Such large datasets usually require lots of effort for their creation, and some tasks, such as medical tasks \cite{prabhu2019few,feyjie2020semi}, may not inherently have enough supervising signals. For these tasks, we need a new paradigm that allows learning a model with a small number of labeled images. The popular FSL tasks \cite{vinyals2016matching,snell2017prototypical,mishra2017simple} serve as a testbed for some certain aspects of such small tasks. Recent efforts toward FSL are summarized as follows.

\begin{figure*}[!t]
	\centering
	\includegraphics[width=\textwidth]{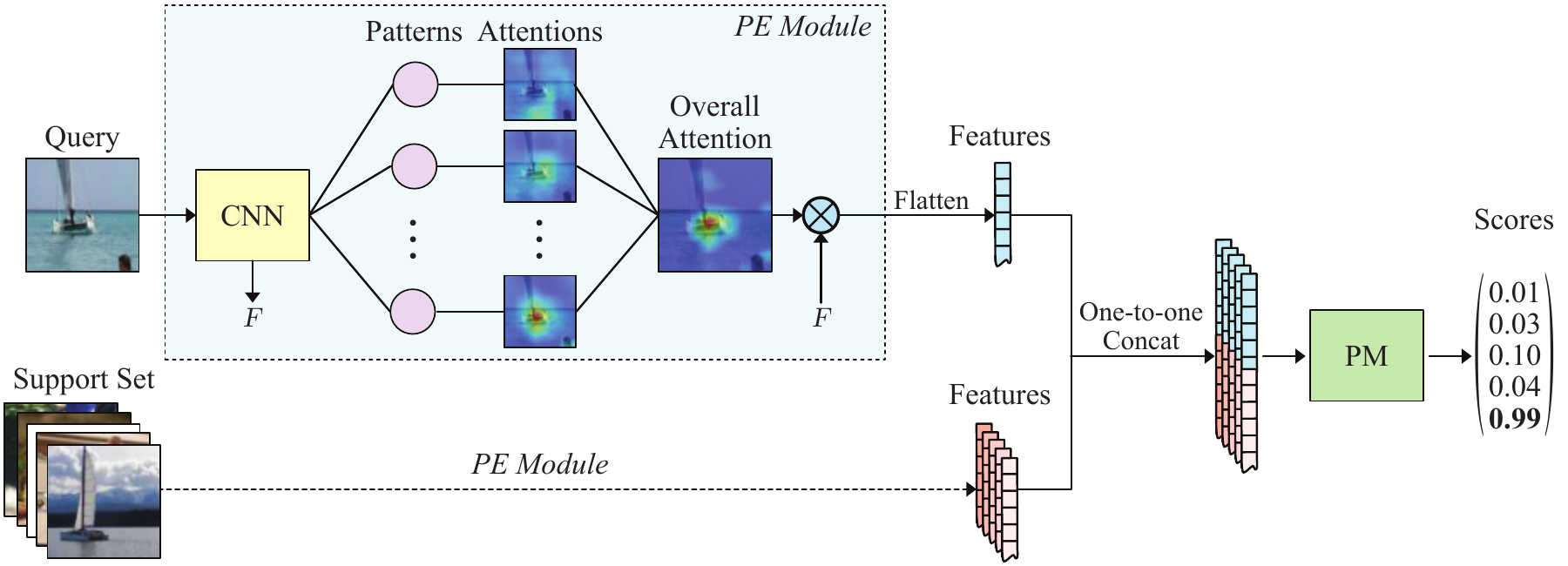}
	\caption{Overall structure of MTUNet. One query is processed by CNN backbone and \textit{pattern extractor} (PE) to provide exclusive patterns and then turned into an overall attention. Query will be concatenated to each support to make a pair for final discrimination through pairwise matching (PM). The dotted line represent each support image undergo the same calculation as query.}
	\label{fig_structure}
\end{figure*}

\textbf{Image Embedding and Metric Learning.}
Many works focus on how to transform images into vectors in embedding space, in which the distance between a pair of vectors represents the conceptual dissimilarity. An early approach uses Siamese networks \cite{koch2015siamese} as a shared feature extractor to produce image embeddings for both support and query images. The weighted $\ell_1$ distance is used for the classification criterion. Some use a multi-layer perceptron (MLP) to parameterize and learn a classifier \cite{kim2019edge, garcia2017few, sung2018learning}. Metric learning can offer a better way to train the mapping into the embedding space \cite{vinyals2016matching,snell2017prototypical}. Some works try to improve the discriminatory power of image embeddings. Simple Shot \cite{wang2019simpleshot} applies $\ell_2$ normalization and Central method to make the distance calculation easier.

\textbf{Meta-learning.}
Another major approach to FSL is to optimize models so that they can rapidly adapt to new tasks. It is a good thing that adapting feature extractor to new tasks at the novel test time. Fine-tuning transfer-learned networks \cite{yosinski2014transferable} fine-tune the feature extractor using the task-specific support images. 
MAML \cite{finn2017model} and its extensions \cite{ravi2016optimization, nichol2018first} train a set of initialization parameters, and through one or more steps of gradient adjustment on the basis of the initial parameters, they can be easily adapted to a new task with only a small amount of data. Besides training a good parameter initialization, Meta-SGD \cite{li2017meta} also trains the parameter update direction and step size.

\textbf{Data Augmentation.}
Data augmentation aims at introducing immutability to models to capture in both image and feature levels \cite{ren2018meta, chen2019image}. Some works try to use samples which are weakly labeled or unlabeled \cite{douze2018low, pfister2014domain}. ICI \cite{wang2020instance} introduces a judgment mechanism to increase training samples. It is always worth increasing the training set by utilizing the unlabeled data with confidently predicted labels. In general, solving an FSL problem by augmenting $D_{train}$ is straightforward and easy to understand.

\textbf{Transductive or Semi-supervised Paradigm.} Transductive or semi-supervised approaches \cite{hu2020leveraging,dhillon2019baseline} have gained popularity, which makes great progress in the past few years. They use the statistics of query examples or statistics across the few-shot tasks, which assumes that all novel images are ready beforehand. We only employ the original inductive paradigm to explore explainable feature extraction, but our idea can be easily adopted to the transductive paradigm.

\subsection{Explainable AI}
Deep neural networks are considered as a black-box technology, and explainable artificial intelligence (XAI) is a series of attempts to unveil them. Most XAI methods for classification tasks are based on back-propagation \cite{Score-CAM,shrikumar2017learning,selvaraju2017grad,chattopadhay2018grad} or perturbation \cite{IBA}. All these methods are \textit{post-hoc}, which can not be added to the model structure during training.

A few works \cite{sunexplain,gengexplainable,karlinsky2020starnet} have tried XAI for FSL tasks. Geng \etal \cite{gengexplainable} uses a knowledge graph to make an explanation for zero-shot tasks. Sun \etal \cite{sunexplain} adopts layer-wise relevance propagation (LRP) \cite{bach2015pixel} to explain the output of a classifier. StarNet \cite{karlinsky2020starnet} realize visualization through heat maps derived from back-project.  Recently, a new type of XAI, coined SCOUTER \cite{li2020scouter}, has been proposed, which applies the self-attention mechanism \cite{Transformer} to the classifier. This method can further extract the discriminant attentions for each category during training, which makes classification results explainable. We apply this technique to FSL tasks in order to explore a new explainable FSL paradigm. PE provides insights on why it classified an input image into a certain novel category. 

\section{Methodology}
\subsection{Problem Definition}\label{definition}
This paper addresses an inductive FSL task (\cf, transductive one \cite{dhillon2019baseline,hu2020leveraging}), in which we are given two disjoint sets  $\mathcal{D}_\text{base}$ and $\mathcal{D}_\text{novel}$ of samples. The former is the base set that includes categories ($\mathcal{C}_\text{base}$) with many labeled images. The latter is the novel set and include categories ($\mathcal{C}_\text{novel}$) with a few labeled images. $\mathcal{C}_\text{base}$ and $\mathcal{C}_\text{novel}$ are disjoint. The FSL task is to find a mapping from a novel image $x$ into the corresponding category $y$.

The literature typically uses the $K$-way $N$-shot episodic paradigm for training/evaluating FSL models. For each episode, we sample two subsets of $\mathcal{D}_\text{base}$ for training, namely, \textit{support set} $\mathcal{S} = \{(x_i, y_i)|i=1,\dots,K \times N\}$ and \textit{query set} $\mathcal{Q} = \{(x_i^\text{q}, y^\text{q}_i)|i=1,\dots,K\times M\}. $ 
These images are of the same $K$ categories in $\mathcal{C}_\text{base}$, and we sampled the same numbers of images ($N$ images for the support set and $M$ images for the query set). An FSL model is trained so that it can find a match between images in $\mathcal{Q}$ (with abuse of notation) and $\mathcal{S}$. The image in $\mathcal{Q}$ is classified as the category of the matched image in $\mathcal{S}$.


\subsection{Overview}

The overall process is illustrated in Figure \ref{fig_structure}. In each episode, we extract feature maps $F=f_\theta(x)\in\mathbb{R}^{c \times h\times w}$ from image $x$ in both $\mathcal{S}$ and $\mathcal{Q}$ using backbone convolutional neural network $f_\theta$, where $\theta$ is the set of learnable parameters. $F$ is then fed into the \textit{pattern extractor} (PE) module, $f_\phi$, with learnable parameter set $\phi$. This module gives attention $A = f_\phi(F) \in \mathbb{R}^{z \times l}$ over $F$. Our \textit{pair matching} (PM) module uses an MLP to compute the score of query image $x^q \in \mathcal{Q}$ belonging to the category of $x$'s in $\mathcal{S}$. 

PE plays a major role in the FSL task. PE is designed to learn a transferable attention mechanism. This ends up in finding common patterns that are shared among different episodes sampled from $\mathcal{D}_\text{base}$. Consequently the patterns are shared also among $\mathcal{D}_\text{novel}$ given that $\mathcal{D}_\text{base}$ and $\mathcal{D}_\text{novel}$ are from similar domains. 


\subsection{Pattern Extractor}

SCOUTER is originally designed as an explainable classifier, of which decision is based directly on the existence of certain learned \textit{patterns} in an image \cite{li2020scouter}. It is built upon the self-attention mechanism \cite{Transformer} to efficiently find common patterns in images of a certain category. In the context of FSL, we extend this idea to find common patterns to efficiently differentiate given sets of categories even across different episodes. The presence of a certain combination of these learned patterns gives a strong clue on the category of the image, facilitating classification even of novel categories. We implement our PE on top of SCOUTER. 

The basic idea of PE is to find common patterns through the self-attention mechanism. Input feature maps $F$ is firstly fed into a $1\times 1$ convolution layer followed ReLU nonlinearity to squeeze the dimensionality of $F$ from $c$ to $d$. We flatten the spatial dimensions of the squeezed features to form $F' \in \mathbb{R}^{d \times l}$, where $l = hw$. To maintain the spatial information, position embedding $P$ \cite{xie2017aggregated, locatello2020object, li2020scouter} is added to the features, \ie, $\tilde{F} = F' + P$.

The self-attention mechanism gives the attention over $F$ for the spatial dimension by the dot-product similarity between a set of $z$ learned patterns $W \in \mathbb{R}^{z \times d}$ ($z$ is the number of the patterns) and $\tilde{F}$ after nonlinear transformations $g_\text{Q}$ and $g_{\text{K}}$. PE repeats this process with updating the pattern with a gated recurrent unit (GRU) to refine the attention. That is, given 
\begin{equation}
    g_\text{Q}(W^{(t)}) \in \mathbb{R}^{z \times d}, \quad g_\text{K}(\tilde{F}) \in \mathbb{R}^{d \times l},
\end{equation}
for the $t$-th repetition, the attention is given using certain normalization function $\xi$ by
\begin{align}
    \bar{A}^{(t)} &= g_\text{Q}(W^{(t)}) g_\text{K}(\tilde{F}) \quad \in (0,1)^{z \times l} \\
    A^{(t)} &= \xi(\bar{A}^{(t)}).
\end{align}
Patterns $W^{(t)}$ is updated $T$ times (\ie, $t=1,\dots,T$) by 
\begin{align}
    U^{(t)} &= A^{(t)} {F'}^\top \\
    W^{(t+1)}&=\operatorname{GRU}{(U^{(t)}, W^{(t)})}.
\end{align}



PE adopts a different normalization strategy from the original SCOUTER. Let $\text{Softmax}_\text{R}(X)$ and $\sigma(X)$ be softmax over respective row vectors of matrix $X$ and sigmoid. SCOUTER normalizes the attention map only over the flattened spatial dimensions, \ie,
\begin{equation}
    A^{(t)} = \sigma(\bar{A}^{(t)}).
\end{equation}
This allows finding multiple patterns in a single image. \Mname{} further modulates this map by 
\begin{equation}
    A^{(t)} =  \sigma(\bar A^{(t)}) \odot  \text{Softmax}_\text{R}(\bar A^{(t)}), \label{eq:att_mod}
\end{equation}
which suppresses weak attention over different patterns at the same spatial location, where $\odot$ is the Hadamard product. This enforces the network to find more specific yet discriminative patterns with fewer correlations among them and thus ends up with more pinpoint attentions. The learned patterns can be more responsive in different images with this modulation as an attention map only responds to a single pattern that does not include its peripheral region.

The input image is finally described by the overall attention corresponding to the extracted patterns, given by
\begin{equation}
    V = \frac{1}{z}A^{(T)} F \mathbf{1}_z,
\end{equation}
where $\mathbf{1}_z$ is the row vector with all $z$ elements being 1. $A^{(T)}$ is reshaped the $l$ into the same spatial structure as $F$. $V$ will then undergo an average pooling among spatial dimension and only keep the channel dimension $c$. 

\subsection{Pairwise Matching}

An FSL classification can be solved by finding the membership of the query to one of the given support images. Some FSL methods use metric learning \cite{snell2017prototypical, vinyals2016matching} to find matches between the query and the supports. The cosine similarity or the $\ell_2$ distance are the typical choices \cite{gidaris2018dynamic,wang2019simpleshot}. Learnable distances are another popular choice for the metric learning-based FSL methods \cite{kim2019edge, garcia2017few, sung2018learning}. We use a learnable distance with an MLP. 

Let $V^\text{q}$ be features of query image $x^\text{q} \in \mathcal{Q}$ and $V_{kn}$ of support image $x_{kn} \in \mathcal{S}$, where the subscripts $k$ and $n$ stand for the $n$-th image of category $k$. For $n > 1$, the average over the $n$ images are taken to generate representative feature $\bar{V}_k$; otherwise (\ie, $n=1$), $\bar{V}_k = V_{k1}$. For computing the membership score $s$ of query image $x^\text{q}$ to category $k$, we use MLP $f_\gamma$ with learnable parameters $\gamma$:
\begin{equation}
s(x^\text{q}, \mathcal{S}_k) = \sigma(f_\gamma([V^\text{q}, \bar{V}_k])), \label{eq:cls}
\end{equation}
where $[\cdot, \cdot]$ is concatenation of two vectors for the  one-to-one pair and $\mathcal{S}_k \subset \mathcal{S}$ contains images of category $k$. $x^\text{q}$ is classified into the category with maximum $s$ over $\mathcal{S}_k$ for $k = 1, 2, \dots, K$.

\subsection{Training}\label{training}
After pre-training of the backbone CNN $f_\theta$, we first train the PE module in the same way as \cite{li2020scouter}. The area loss facilitates to find compact patterns. For this training, we further split $\mathcal{D}_\text{base}$ into two subsets $\mathcal{D}_\text{base,T}$ and $\mathcal{D}_\text{base,V}$. The former contains 90\% of images of each category and the latter contains the rest. We sample $z$ categories from $\mathcal{D}_\text{base}$ and use images of these categories in $\mathcal{D}_\text{base,T}$ for training. The images of the same sampled categories in $\mathcal{D}_\text{base,V}$ is used for validation. With these sampled categories, the training is trying to find discriminative patterns together with our attention map modulation in Eq.~(\ref{eq:att_mod}).

We then train \mname{} with the backbone and the PE module fixed. For $\mathcal{Q}$ and $\mathcal{S}$ sampled from $\mathcal{D}_\text{base}$ for each episode, we train $f_\gamma$ with the binary cross-entropy loss:
\begin{equation}
    L = -\sum_{(x^\text{q}, y^\text{q}) \in \mathcal{Q}}  {y^\text{q}}^\top \log(s(x^\text{q},\mathcal{S})),
\end{equation}
where $s(x^\text{q},\mathcal{S}) = (s_1(x^\text{q},\mathcal{S}_1) ,  \dots, s_K(x^\text{q},\mathcal{S}_K) )^\top$.


\section{Experiments}
\subsection{Datesets}
We evaluate our approach on three commonly-used datasets, mini-ImageNet \cite{vinyals2016matching}, tiered-ImageNet \cite{ren2018meta}, and CIFAR-FS \cite{bertinetto2018meta}. \textbf{Mini-ImageNet} consists of 100 categories sampled from ImageNet with 600 images per class. These images are divided into the base $\mathcal{D}_\text{base}$, novel validation $\mathcal{D}_\text{val}$, and novel test $\mathcal{D}_\text{test}$ sets with 64, 16, and 20 categories, respectively, where both $\mathcal{D}_\text{val}$ and $\mathcal{D}_\text{test}$ corresponded to $\mathcal{D}_\text{novel}$ in Section \ref{definition}. The images in miniImageNet are of size $84\times84$. \textbf{Tiered-ImageNet} consists of ImageNet 608 classes divided into 351 base classes, 97 novel validation classes, and 160 novel test classes. There are 779,165 images with size $84\times84$. \textbf{CIFAR-FS} is a dataset with images from CIFAR-100 \cite{krizhevsky2009learning}. This dataset contains 100 categories with 600 images each. We follow the split given in \cite{bertinetto2018meta}, which are 64, 16, and 20 categories for the base, novel validation, and novel test sets.

\subsection{Experimental Setup}\label{experiment}

Following the majority of the literature, we evaluate \mname{} on 10,000 episodes of 5-way classification created by first randomly sampling 5 categories from $\mathcal{D}_\text{base}$ and then sampling support and query images of these categories with $N = 1$ or $5$ and $M = 15$ per category. We report the average accuracy over $K\times M = 75$ queries in the 10,000 episodes and the 95\% confidence interval.

We employ two CNN architectures as our backbone $f_\theta$, which are often used for FSL tasks, namely WRN-28-10 \cite{zagoruyko2016wide} and ResNet-18 \cite{ResNet}. 
For ResNet-18, we remove the first two down-sampling layers and change the kernel of the first $7\times7$ convolutional layer to $3\times3$. We use the hidden vector of the last convolutional layer after ReLU as feature maps $F$, where the numbers of feature maps are 512 and 640 for ResNet-18 and WRN-28-10 respectively.

\renewcommand{\thefootnote}{\fnsymbol{footnote}}
\setcounter{footnote}{2}
\footnotetext{Results are reported in \cite{wang2019simpleshot}}
\setcounter{footnote}{2}

\begin{table}[!t]
\caption{Average accuracy of 10000 sampling 5-ways task on test set of mini-ImageNet.}
\centering
\resizebox{0.95\columnwidth}{!}{
\begin{tabular}{lcc}
\toprule
\textbf{Approach} &\textbf{One shot} &\textbf{Five shots}\\
\midrule
MetaLSTM \cite{ravi2016optimization} &43.44$\pm$0.77 &60.60$\pm$0.71 \\
MatchingNet \cite{vinyals2016matching} &43.56$\pm$0.84 &55.31$\pm$0.73 \\
MAML \cite{finn2017model} &48.70$\pm$1.84 &63.11$\pm$0.92\\
LLAMA \cite{grant2018recasting} &49.40$\pm$1.83 &-\\
ProtoNet \cite{snell2017prototypical} &49.42$\pm$0.78 &68.20$\pm$0.66\\
PLATIPUS \cite{finn2018probabilistic} &50.13$\pm$1.86 &-\\
GNN \cite{garcia2017few} &50.33$\pm$0.36 &66.41$\pm$0.63\\
RelationNet \cite{sung2018learning} &50.44$\pm$0.82 &65.32$\pm$0.70\\
Meta SGD \cite{li2017meta} &50.47$\pm$1.87 &64.03$\pm$0.94\\
R2-D2 \cite{bertinetto2018meta} &51.20$\pm$0.60 &68.20$\pm$0.60\\
RelationNet \cite{sung2018learning} &52.48$\pm$0.86 &69.83$\pm$0.68\\
Gidaris \cite{gidaris2018dynamic} &55.45$\pm$0.89 &70.13$\pm$0.68\\
SNAIL \cite{mishra2017simple} &55.71$\pm$0.99 &68.88$\pm$0.92\\
adaCNN \cite{munkhdalai2018metalearning} &56.88$\pm$0.62 &71.94$\pm$0.57\\
SimpleShot(UN) \cite{wang2019simpleshot} &57.81$\pm$0.21 &\textbf{80.43$\pm$0.15}\\
Qiao \cite{qiao2018few} &59.60$\pm$0.41 &73.74$\pm$0.19\\
LEO \cite{rusu2018meta} &\textbf{61.76$\pm$0.08} &77.59$\pm$0.12\\
\midrule
\mname{}+ResNet-18 &55.03$\pm$0.49 &70.22$\pm$0.35\\
\mname{}+WRN &56.12$\pm$0.43 &71.93$\pm$0.40\\
\bottomrule
\end{tabular}
}
\label{results_mini}
\end{table}

\begin{table}[!t]
\caption{Average accuracy of 10000 sampling 5-ways task on test set of tiered-ImageNet.}
\centering
\resizebox{0.95\columnwidth}{!}{
\begin{tabular}{lcc}
\toprule
\textbf{Approach} &\textbf{One shot} &\textbf{Five shots}\\
\midrule
Reptile \cite{nichol2018first}\footnotemark[\value{footnote}] &48.97$\pm$0.21 &66.47$\pm$0.21\\
MAML \cite{finn2017model} &51.67$\pm$1.81 &70.30$\pm$0.08\\
ProtoNet \cite{snell2017prototypical}\footnotemark[\value{footnote}] &53.31$\pm$0.20 &72.69$\pm$0.74\\
RelationNet \cite{sung2018learning} &54.48$\pm$0.93 &71.32$\pm$0.78\\
Meta SGD \cite{li2017meta}\footnotemark[\value{footnote}] &62.95$\pm$0.03 &79.34$\pm$0.06\\
SimpleShot(UN) \cite{wang2019simpleshot} &64.35$\pm$0.23 &\textbf{85.69$\pm$0.15}\\
LEO \cite{rusu2018meta} &\textbf{66.33$\pm$0.05} &81.44$\pm$0.09\\
\midrule
\mname{}+ResNet-18 &61.27$\pm$0.50 &77.82$\pm$0.41\\
\mname{}+WRN &62.42$\pm$0.51 &80.05$\pm$0.46\\
\bottomrule
\end{tabular}
}
\label{results_tiered}
\end{table}

\setcounter{footnote}{3}
\footnotetext{Results are reported in \cite{bertinetto2018meta}}
\setcounter{footnote}{3}

\begin{table}[!t]
\caption{Average accuracy of 10000 sampling 5-ways task on test set of CIFAR-FS.}
\centering
\resizebox{0.95\columnwidth}{!}{
\begin{tabular}{lcc}
\toprule
\textbf{Approach} &\textbf{One shot} &\textbf{Five shots}\\
\midrule
RelationNet \cite{sung2018learning}\footnotemark[\value{footnote}] &55.00$\pm$1.00 &69.30$\pm$0.80\\
ProtoNet \cite{snell2017prototypical}\footnotemark[\value{footnote}] &55.50$\pm$0.70 &72.00$\pm$0.60\\
MAML \cite{finn2017model}\footnotemark[\value{footnote}] &58.90$\pm$1.90 &71.50$\pm$1.00\\
GNN \cite{garcia2017few}\footnotemark[\value{footnote}] &61.90 &75.30\\
R2-D2 \cite{bertinetto2018meta} &65.30$\pm$0.20 &78.30$\pm$0.20\\
\midrule
\mname{}+ResNet-18 &66.31$\pm$0.50 &80.16$\pm$0.39\\
\mname{}+WRN &\textbf{68.34$\pm$0.49} &\textbf{82.93$\pm$0.37}\\
\bottomrule
\end{tabular}
}
\label{results_CIFAR-FS}
\end{table}

\begin{figure*}[!t]
	\centering
	\includegraphics[width=\textwidth]{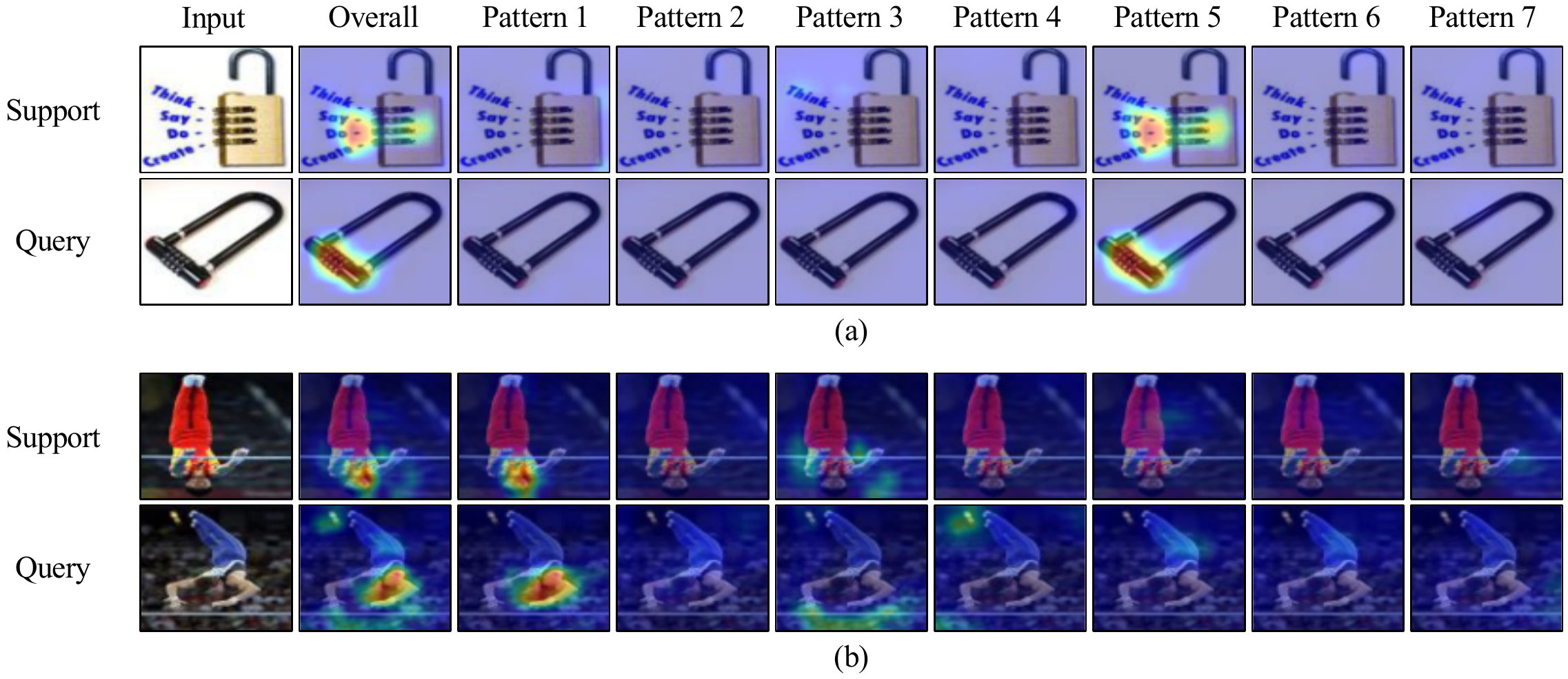}
	\caption{Visualization of each pattern and the average features for a sampled task in mini-ImageNet. a is the class of lock and c is the horizontal bar. Overall is the overall attention among all patterns. The third to ninth columns are the visualization of the re-gions  corresponding  to  the  learned  patterns.}
	\label{fig:xslot}
\end{figure*}

As noted in \cite{li2020scouter}, pre-training of the backbone CNNs is important for our PE module. We adopted a distance-based strategy, which is similar to SimpleShot \cite{wang2019simpleshot}. We train the backbone CNNs with all images in $\mathcal{D}_\text{base}$. The performance of a simple nearest-neighbor-based method is then evaluated over $\mathcal{D}_\text{val}$ with 2,000 episodes of 5-way FSL task, and the best model is adopted. The learning rate for training starts with $10^{-3}$ and is divided by 10 every 20 epochs. It has total of 50 epochs.

As for the PE module, we set $d$ to 64. We set the maximum update $T$ as 3. For the number $z$ of the patterns is empirically set to $1/10$ of the number of the base set categories, which are 7, 36, and 7 for mini-ImageNet, tiered-ImageNet, and CIFAR-FS, respectively. The importance of this choice is discussed in Section \ref{sec:discussion}. SCOUTER's loss has two hyper-parameters $e$ and $\lambda$, which controls over positive/negative explanation and over the preference to smaller attention areas, respectively. We set $e=1$ (\ie, positive explanation) and $\lambda=1$ following \cite{li2020scouter}. Both $g_\text{Q}$ and $g_\text{K}$ have three FC layers with ReLU nonlinearities between them. All the parameters in the backbone $f_\theta$ are fixed. We adopt the training strategy described in \ref{training}. The learning rate for training starts with $10^{-4}$ and is divided by 10 every 40 epochs. It is trained for total of 60 epochs.

For training the whole MTUNet, the learnable parameters in backbone CNNs and PE are frozen. 
In a single epoch of training, we sample 1,000 episodes of 5-way tasks. The model is trained for 20 epochs with an initial learning rate $10^{-3}$, which is divided by 10 at the 10-th epoch. We use the model with the best performance with 2,000 episodes sampled from $\mathcal{D}_\text{val}$.

Our model is implemented with PyTorch. AdaBelief \cite{zhuang2020adabelief} is adapted as optimizer. Input images are resized into $80\times80$, going through data augmentation including random flip and affine transformation following \cite{wang2019simpleshot}. A GPU workstation with two NVIDIA Quadro GV100 (32GB memory) GPUs are used for all experiments.

\subsection{Results}\label{sota}
\Mname{} is compared with state-of-the-art (SoTA) FSL methods. We exclude ones in semi-supervised and transductive paradigms, which use the statistics of novel set across different FSL tasks. We also do not adopt any post-processing methods like $\ell_2$ normalization in \cite{wang2019simpleshot}. 

We report our best model by randomly sampling 10,000 1-shot and 5-shot tasks over $\mathcal{D}_\text{test}$ in Tables \ref{results_mini}--\ref{results_CIFAR-FS} for the three datasets. We select the category for PE pre-training by sampling every 10 categories from the base set category list. The results demonstrate that \mname{} outperforms or is comparable with SoTA methods. 

The different architectures of the backbone CNNs affect the performance. The variants with WRN always give a better performance than those with ResNet-18. Asides from the difference in the network architecture, the size of feature maps may be one of the factors. For mini-ImageNet, the WRN variants has $20 \times 20$ feature maps, while the ResNet-18 variants has $10 \times 10$.  Such larger feature maps not only provides more information to the \cname{} module but also give a better basis of patterns as higher resolutions may help find more specific patterns.

\subsection{Explainability}\label{explain}
In addition to the classification performance, \mname{} is designed to be explainable in two different aspects. Firstly, \mname{}'s decision is based on certain combinations of learned patterns. These patterns are localized in both query and support images through $A^{(T)}$, which can be easily visualized. This visualization offers intuition on the learned patterns and how much these patterns are shared among the query and support images. Secondly, thanks to the one-to-one matching strategy formulated as a binary classification problem in Eq.~(\ref{eq:cls}), the distributions (or appearances) of learned patterns in query and support images give a strong clue on \Mname{}'s matching score $s$. 


\paragraph{Pattern-based visual explanation.}
\mname{}'s decision is based on the learned patterns, \ie, it is solely based on how much shared patterns (or features) appear in both a query and a support. This design in turn means that, by pinpointing each pattern in the images, we can obtain an intuition behind the decision made by the model. This can be done by merely visualizing $A^{(T)}$. 

Figures \ref{fig:xslot} (a) and (b) respectively show a pair of support and query images in a 5-way task in mini-ImageNet. The pairs (a) and (b) are of categories \texttt{lock} and \texttt{horizontal bar}, respectively. The second column shows the visualization of averaged attention, given by
\begin{equation}
    A' = \frac{1}{z} A^{(T)} \mathbf{1}_z.
\end{equation}
The third to ninth columns are the visualization of the regions corresponding to the learned patterns in $A^{(T)}$ (\ie, the $i$-th row vectors of $A^{(T)}$ represents the appearance of the $i$-th learned pattern at the respective spatial location).

For (a) with category \texttt{lock}, the support image is a small golden combination lock used for storage cabinets or post boxes. Among all 7 patterns, only pattern 5 shows a strong response, whereas the others are not observed. We can see that pattern 5 pays attention to the discs of the lock. It also gives a strong response at the words on the left which shows similar morphological characteristics. The query image of (a) is a black combination lock often used for bicycles. The attention maps show almost the same distributions as the support: Only pattern 5 has a response on the discs. From these visualizations, we can infer that pattern 5 represents periodical changes in colors. Although these two locks have different functions, \mname{} finds a shared pattern among them. 

For (b), the support image is the gymnast wearing red. Multiple patterns are observed in the image. We can see that the visualization of pattern 1 identifies the part of the human body (head), and pattern 3 appears around the hands grabbing the horizontal bar. The query image is the gymnast in blue. Patterns 1 and 3 respond in a similar way to the support image. Patters 4 and 5 appear in the background and around other parts of the body, however their responses are relatively weak compared to patterns 1 and 3. Patterns 1 and 3 may responsible for human heads and hands grabbing the horizontal bar, lead to the successful classification of novel categories.

\begin{figure}[!t]
	\centering
	\includegraphics[width=.9\columnwidth]{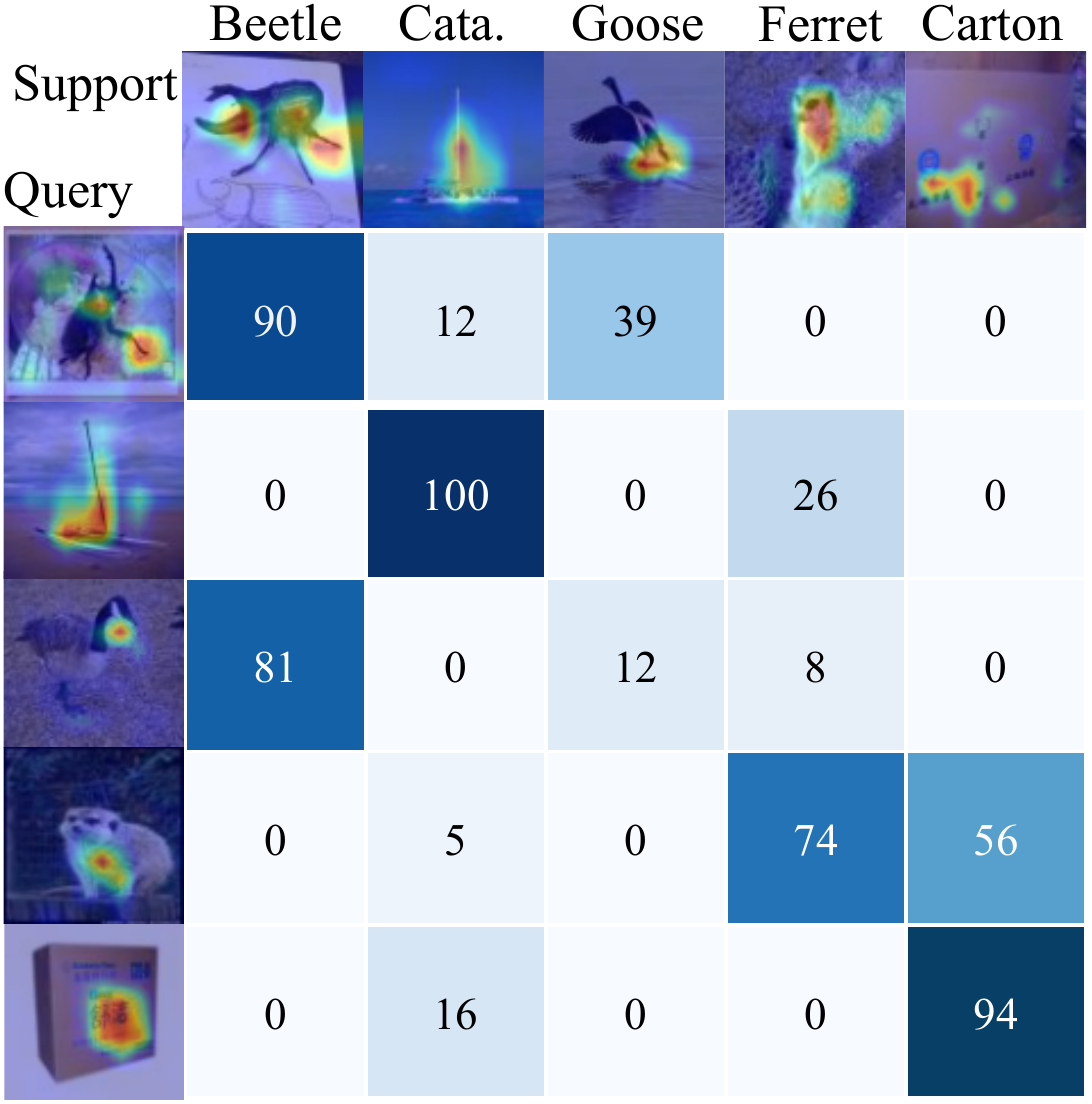}
	\caption{Matching point matrix of one sampled task in mini-ImageNet. Row and column are consisted with the overall attention visualization for support and query of each category.}
	\label{fig:score}
\end{figure}

\paragraph{Visualization of pairwise matching scores}
Figure \ref{fig:score} shows the visualization of the pairwise matching score of a 5-way 1-shot task over mini-ImageNet, compiled in a matrix. Through the pairwise matching module, an FSL task is cast into a binary classification problem. The output for each pair is a value between 0 to 1 due to the sigmoid function, whereas the scores are shown in percentage in the figure. The first row and the first column are the visualization of overall attention for the support and query images of each category.

Among all pairwise combinations, the combination of the support and query images of \texttt{catamaran} makes the full score (100\%). The visualization of overall attention covers the hulls, especially the masts, in both images, which are the main characteristics of this category, explaining the high score. Category \texttt{goose} gets a low matching score. The query is a close-up of a goose on the ground from its front side, which captures the goose's blackhead or beak. The support image is an overall view of a goose about to fly from its backside. The visualization of overall attention captures the leg. With this combination, finding a shared pattern may not be easy, although these two extracted patterns are both representative parts of a bird. This problem stems from the difference in viewing angles, which can be relieved in 5-shot tasks, giving more supports from different viewing angles. Surprisingly, the query image for \texttt{goose} gets 81\% for the support image for \texttt{beetle}. This may suggest that one of the patterns responds to black regions and this pattern is solely used as the clue of \texttt{goose}. This is a negative result for FSL tasks but clearly demonstrates \mname{}'s explainability on the relationship between visual patterns and the pairwise matching scores.

\subsection{Discussion}\label{sec:discussion}

\begin{figure}[!t]
	\centering
	\includegraphics[width=1.0\columnwidth]{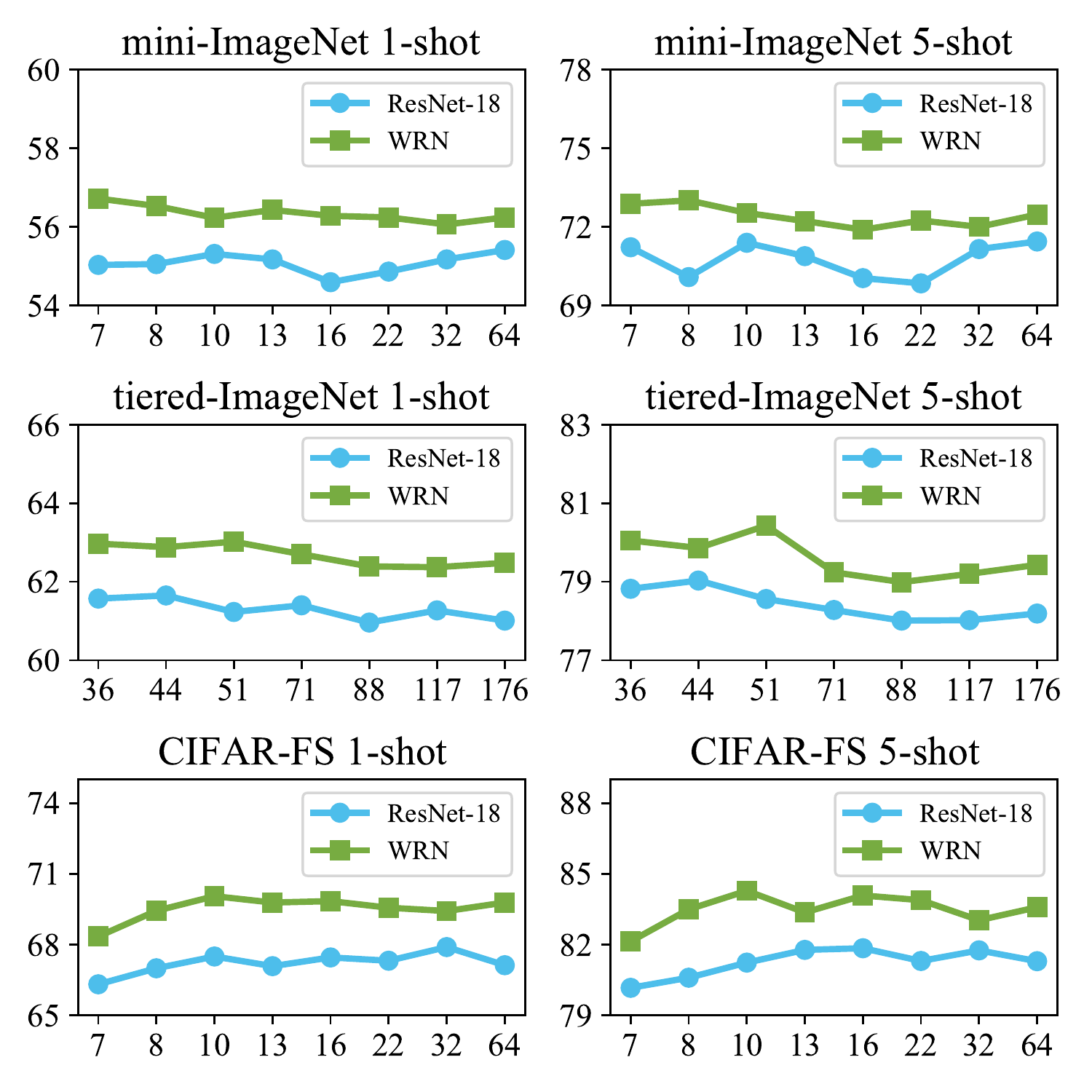}
	\caption{Results of patterns number settings for mini-ImageNet, tiered-ImageNet, and CIFAR-FS. The horizontal axis represents the number of patterns and the vertical axis represents the average accuracy. We report all the results with 10,000 sampled 5-way episodes in the novel test set.}
	\label{fig:ablation}
\end{figure}

\begin{figure}[!t]
	\centering
	\includegraphics[width=0.9\columnwidth]{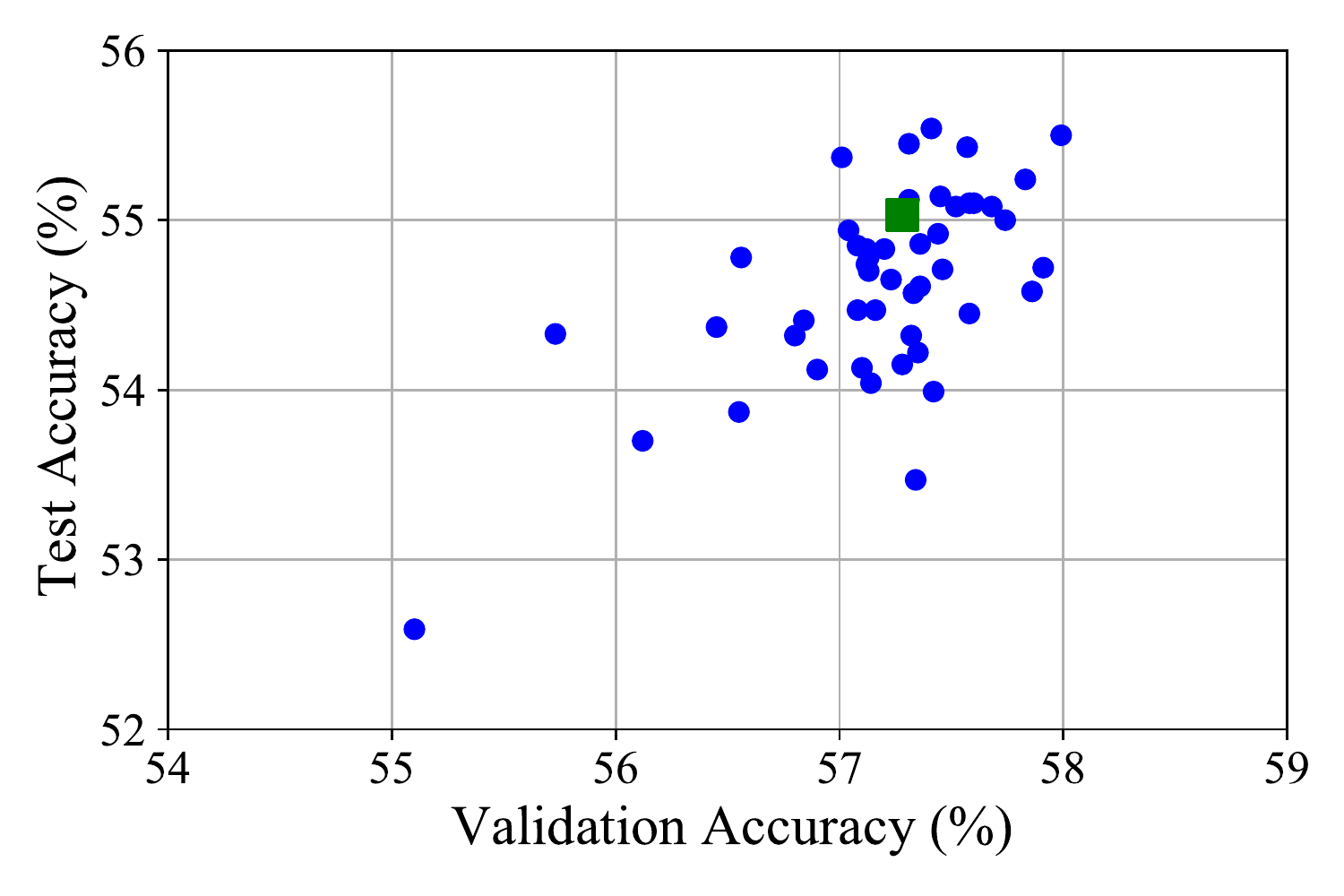}
	\caption{Performance of 50 turns random categories sampling for PE pre-training with 7 patterns. All the experiments are implemented in mini-ImageNet using ResNet-18 as the backbone. Result of the sample for our experiments is marked in green.}
	\label{fig:scatter}
\end{figure}

\paragraph{The number $z$ of patterns.} The number of patterns can be another crucial factor for \mname{}. Intuitively, a larger $z$ makes the model more discriminative. To show the impact of $z$, we uniformly sample categories in $\mathcal{C}_\text{base}$ (\ie, default as sampling every $I$ categories from the category list, where $I=10, 8, 7, 5, 4, 3, 2,$ and $1$); thus, $I=1$ ends up with using all categories in $\mathcal{C}_\text{base}$. 

The test accuracies are shown in Figure \ref{fig:ablation} for 5-way 1-shot and 5-way 5-shot tasks in 10,000 sampled episodes over $\mathcal{D}_\text{test}$ of the three datasets. The horizontal axis represents the number of patterns and the vertical axis represents the average accuracy. Interestingly, the results show no clear tendency with respect to $z$. We would say that the performance is slightly decreased in mini-ImageNet with a larger $z$, whereas slightly increased in CIFAR-FS. For tiered-ImageNet, when setting $I$ as 1 and use all 351 base categories for patterns, the PE module can not be trained successfully. This situation is also reported in \cite{li2020scouter}. Other settings also show no obvious differences. In general, tuning over $z$ may help gain performance, but its impact is not significant. 

\paragraph{Selection of categories for training PE.}
Our PE module is supposed to learn common visual patterns. We use images of a certain subset of categories in $\mathcal{C}_\text{base}$ to learn such patterns in our experiments. The selection of this subset thus affects the performance of downstream FSL tasks. To clarify the impact of the choice of the subset, we randomly sample seven categories in $\mathcal{C}_\text{base}$ of mini-ImageNet for 50 times and use the corresponding images for training PE on top of ResNet-18. The trained PE is used for training \mname{}, which is evaluated over 2,000 episodes of FSL tasks with both the validation and test sets. 

The mean and the 95\% confidence interval over the 50 test accuracies are 54.63\% and 0.16\%, respectively. This implies that our model benefits from a better choice of categories for training PE. For this choice, we only have access to the validation set; however, since the validation set and the test set have disjoint categories, the best choice for the validation set is not necessarily the best choice for the test set. Figure \ref{fig:scatter} shows the scatter plot of the validation accuracies and corresponding test accuracies, over 50 different random samples of seven categories. The plot empirically shows that the validation and test accuracies are highly correlated to each other, with Pearson's correlation coefficient of 0.64. This leads to the conclusion that, at least for mini-ImageNet, we can use the validation set to find the better choice. The green square in the plot is the choice that we adopted in our experiments, which shows that it is a better choice but not the best. 



\section{Conclusion}
In this paper, we propose \mname{} designed for explainable FSL. We achieved comparable performance on three benchmark datasets and qualitatively demonstrated its strong explainability through patterns in images. The approach taken in our model might be analogous to human beings as we usually try to find shared patterns when making a match between images of an object that one has never seen before. This can be advantageous as the explanation given by \mname{} can provide an intuitive interpretation of what the model actually does. Our future work includes testing our model in a practical application scenario of FSL, such as computer-assisted diagnoses. 

\section{Acknowledgements}
This work was supported by Council for Science, Technology and Innovation (CSTI), cross-ministerial Strategic Innovation Promotion Program (SIP), ``Innovative AI Hospital System'' (Funding Agency: National Institute of Biomedical Innovation, Health and Nutrition (NIBIOHN)). This work was also supported by JSPS KAKENHI Grant Number 19K10662 and 20K23343.

{\small
\bibliographystyle{ieee_fullname}
\bibliography{bib}
}
\vspace{6.5in}

\pagebreak
\clearpage

\begin{strip}
\begin{center}
	 {\Large \bf Match Them Up: Visually Explainable Few-shot Image Classification\\(Supplementary Material) \par}
	
	 \vskip .5em
	 \vspace*{12pt}
\end{center}
\vspace{0.5 in}
\end{strip}
\setcounter{equation}{0}
\setcounter{figure}{0}
\setcounter{table}{0}
\setcounter{section}{0}
\setcounter{page}{1}
\makeatletter

\section{Selection of Categories for Training PE.}
We further provide the randomly sampled seven categories in $\mathcal{C}_\text{base}$ of cifar-FS and tiered-ImageNet for 50 times and 20 times respectively. ResNet-18 is used as backbone. \mname{} with the trained PE is evaluated over 2,000 episodes of FSL tasks with both the validation and test sets. 

\begin{figure}[!t]
	\centering
	\includegraphics[width=1\columnwidth]{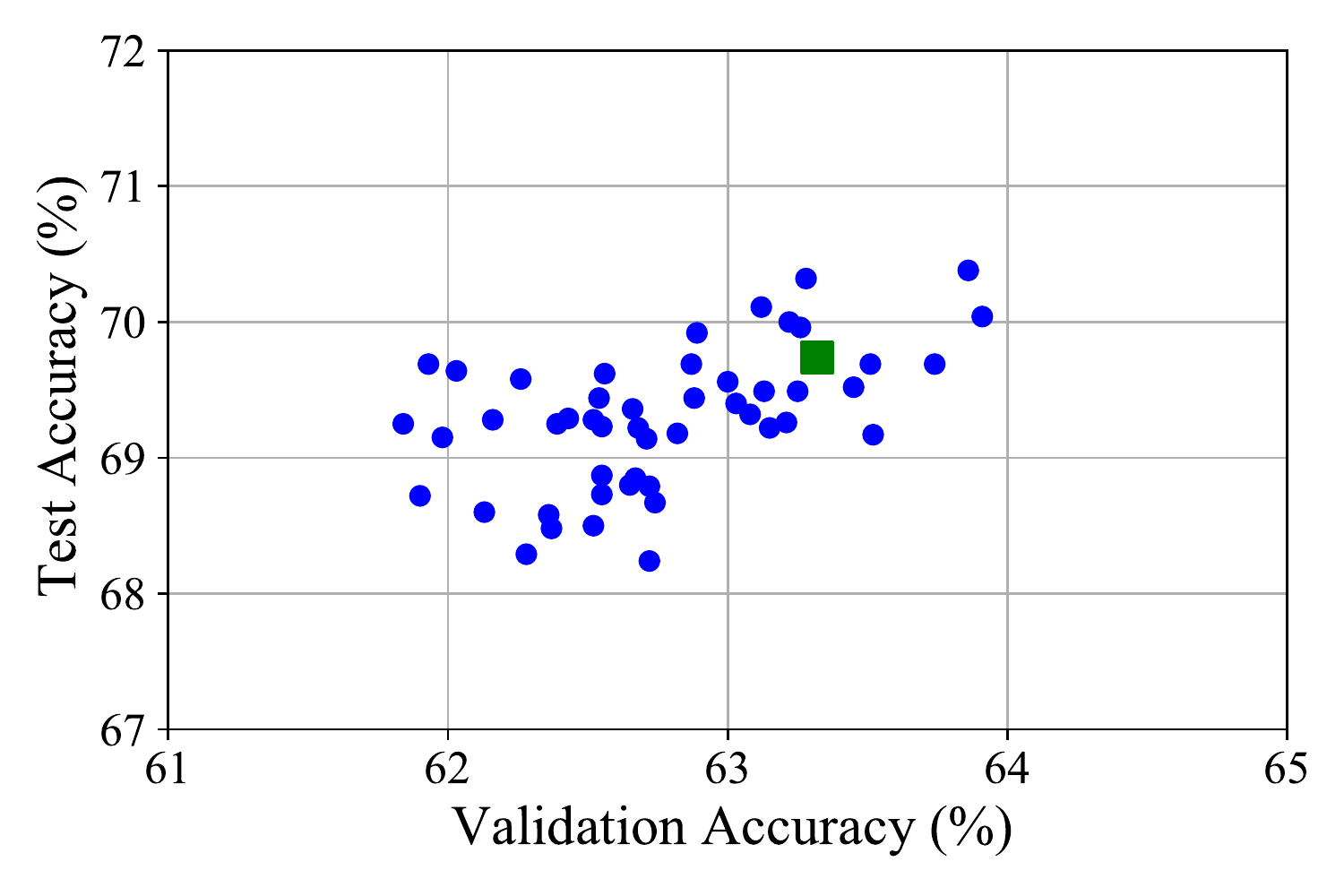}
	\caption{PE training experiments implemented in cifar-FS.}
	\label{cifarfs}
\end{figure}

\begin{figure}[!t]
	\centering
	\includegraphics[width=1\columnwidth]{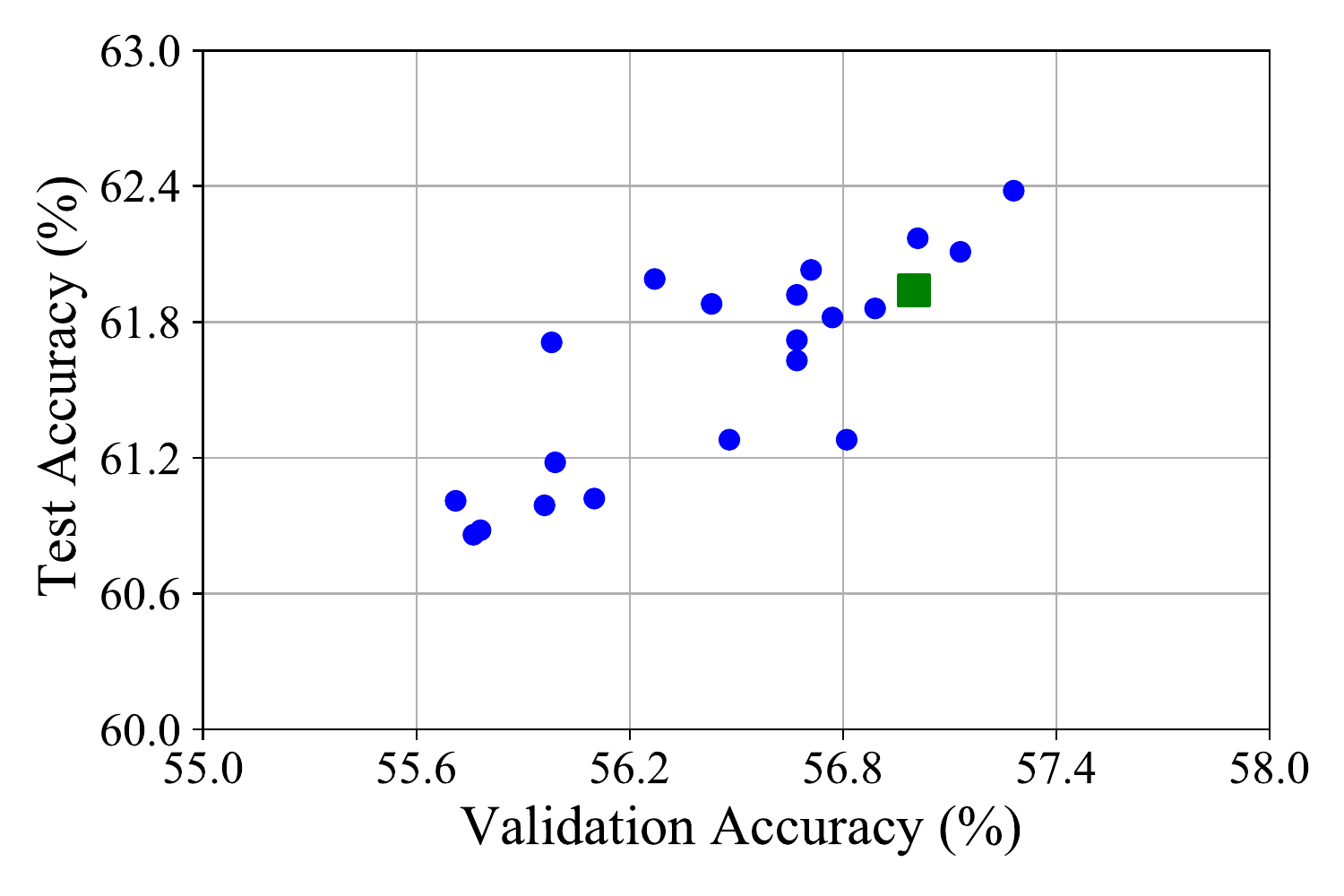}
	\caption{PE training experiments implemented tiered-ImageNet.}
	\label{tiered}
\end{figure}

The results are shown in Figures \ref{cifarfs} and \ref{tiered}. For cifar-FS, the mean and the 95\% confidence interval over the 50 test accuracies are 69.29\% and 0.14\%, respectively. Pearson's correlation coefficient is 0.53. For tiered-ImageNet, the mean and the 95\% confidence interval over the 20 test accuracies are 61.58\% and 0.20\%, respectively. Pearson's correlation coefficient is 0.81. Through the result we can find that the performances over the validation and test sets show a strong correlation. These results imply that we can use the validation set to find a better choice not only for miniImageNet but also the other datasets.

\section{Explainability}
\begin{figure*}[!t]
	\centering
	\includegraphics[width=\textwidth]{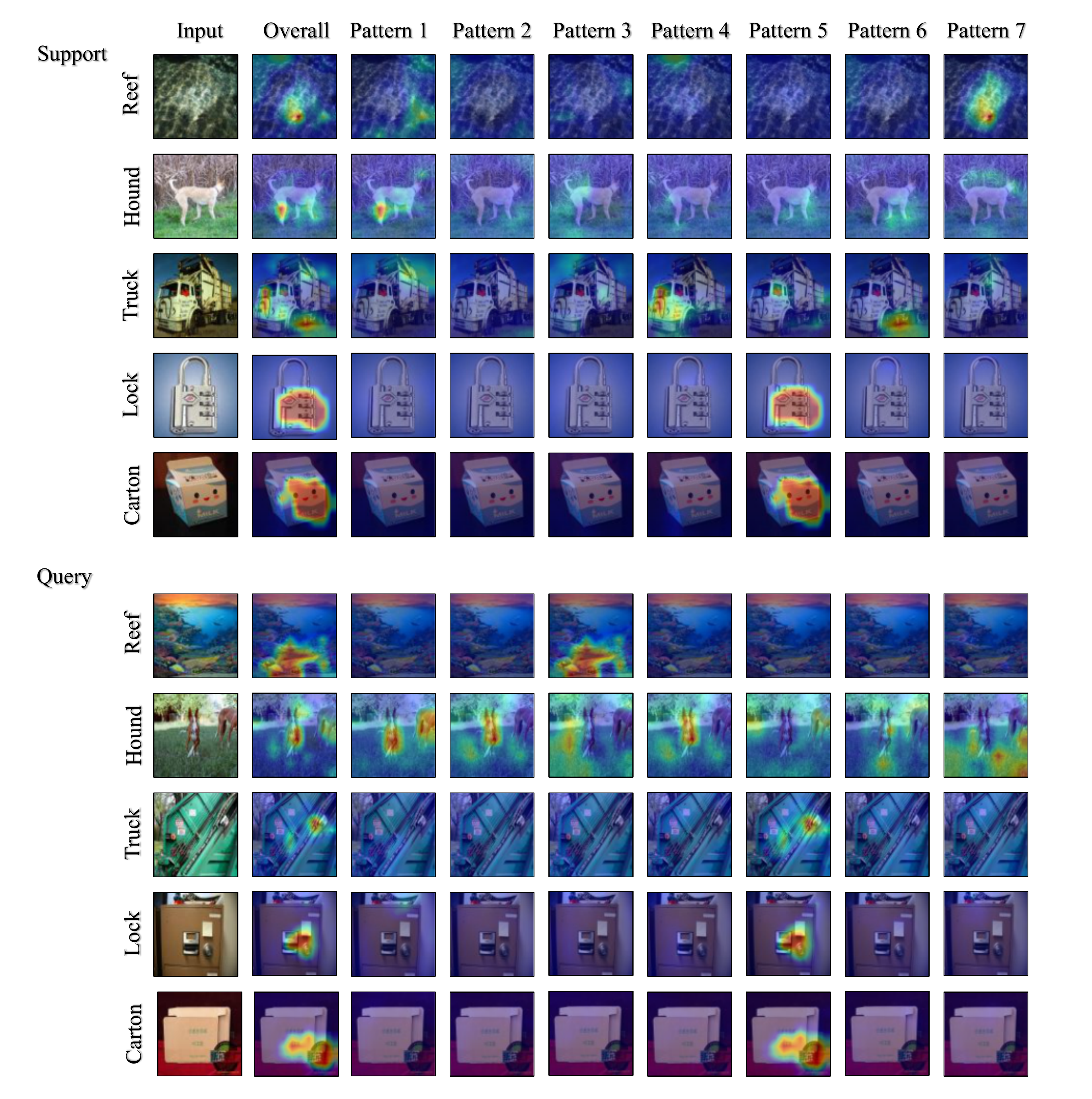}
	\caption{Pattern-based visualization of sample 1.}
	\label{pair1}
\end{figure*}

\begin{figure*}[!t]
	\centering
	\includegraphics[width=\textwidth]{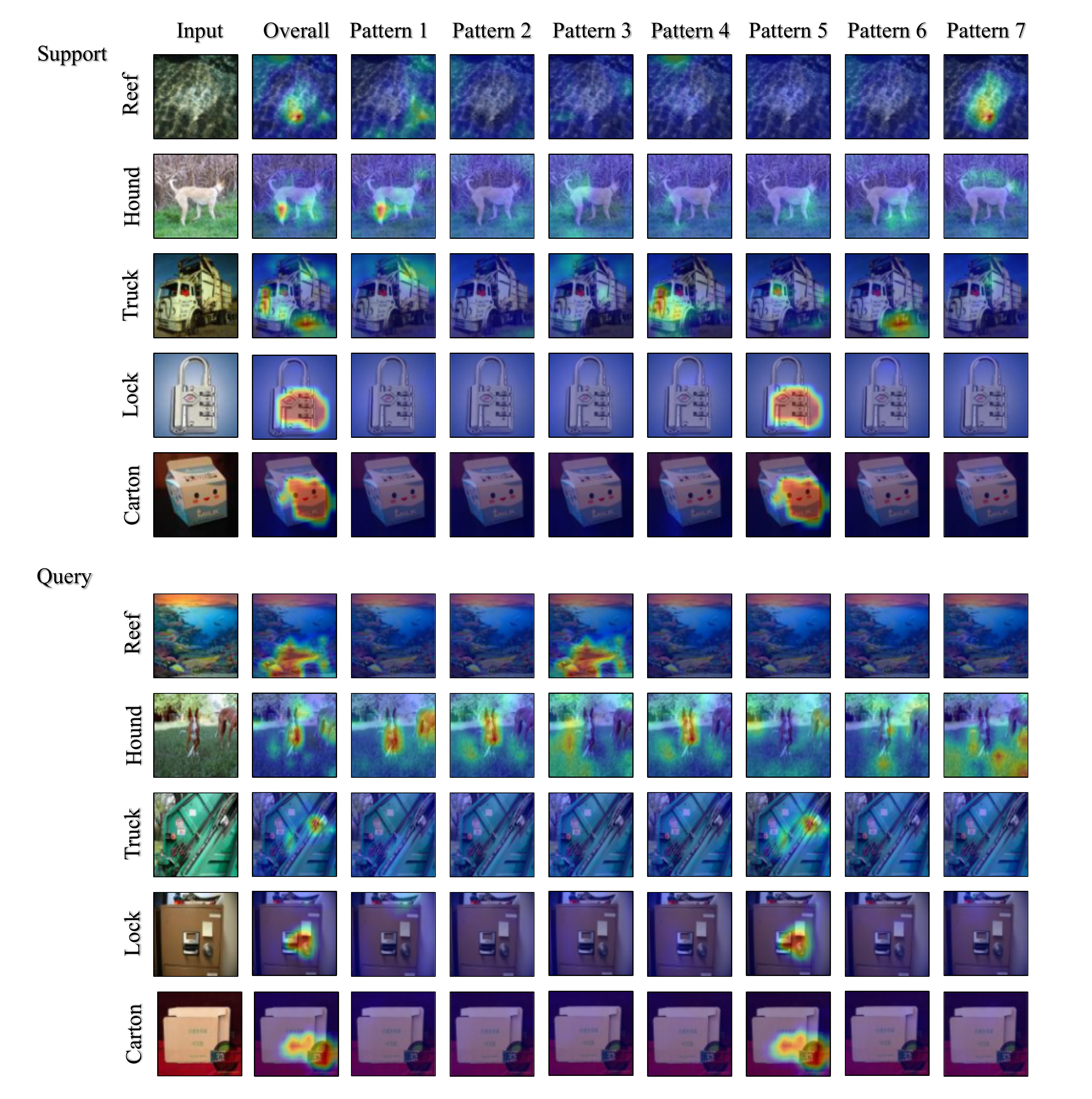}
	\caption{Pattern-based visualization of sample 2.}
	\label{pair2}
\end{figure*}

\begin{figure*}[!t]
	\centering
	\includegraphics[width=\textwidth]{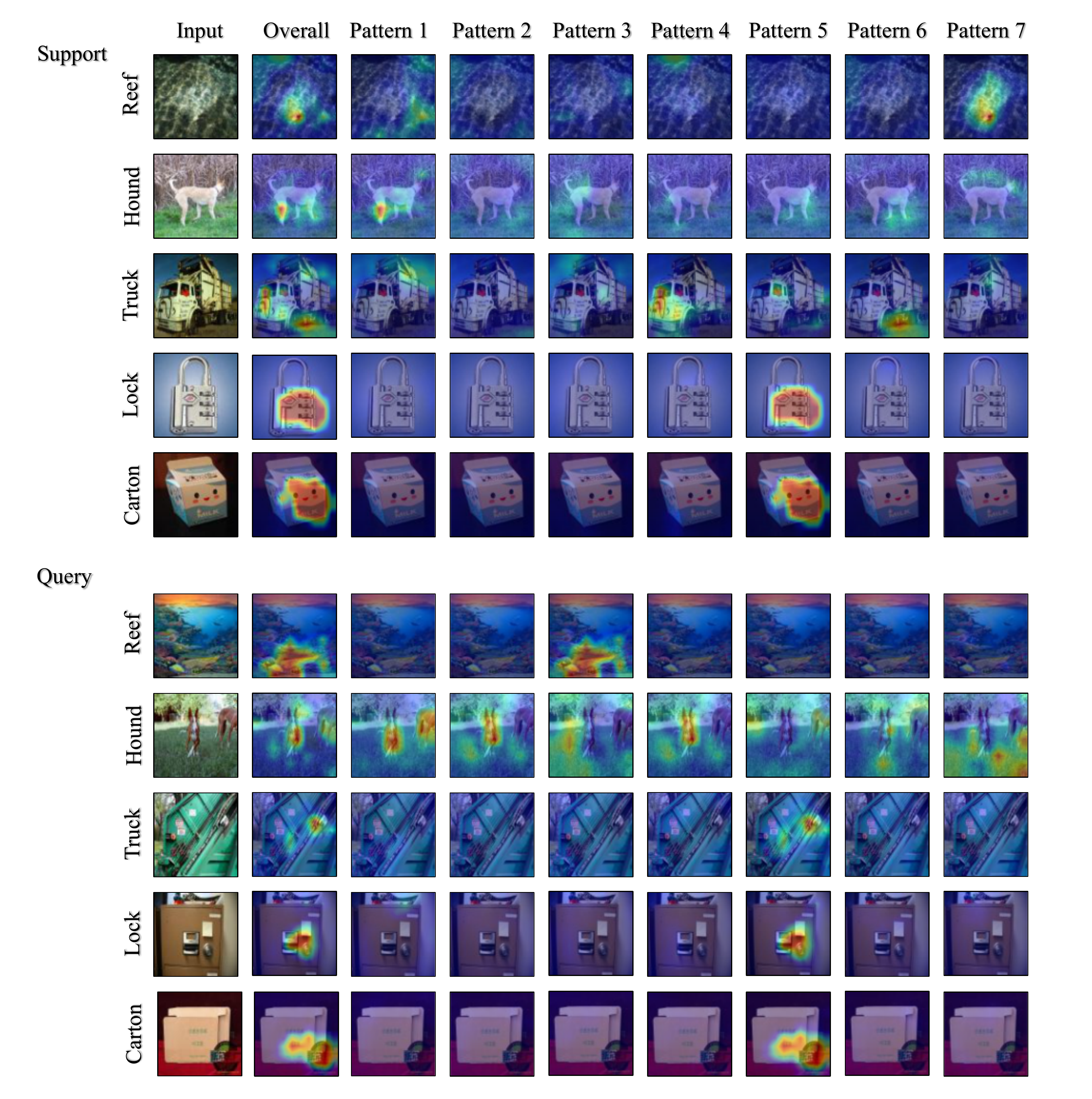}
	\caption{Pattern-based visualization of sample 3.}
	\label{pair3}
\end{figure*}

\begin{figure*}[!t]
	\centering
	\includegraphics[width=\textwidth]{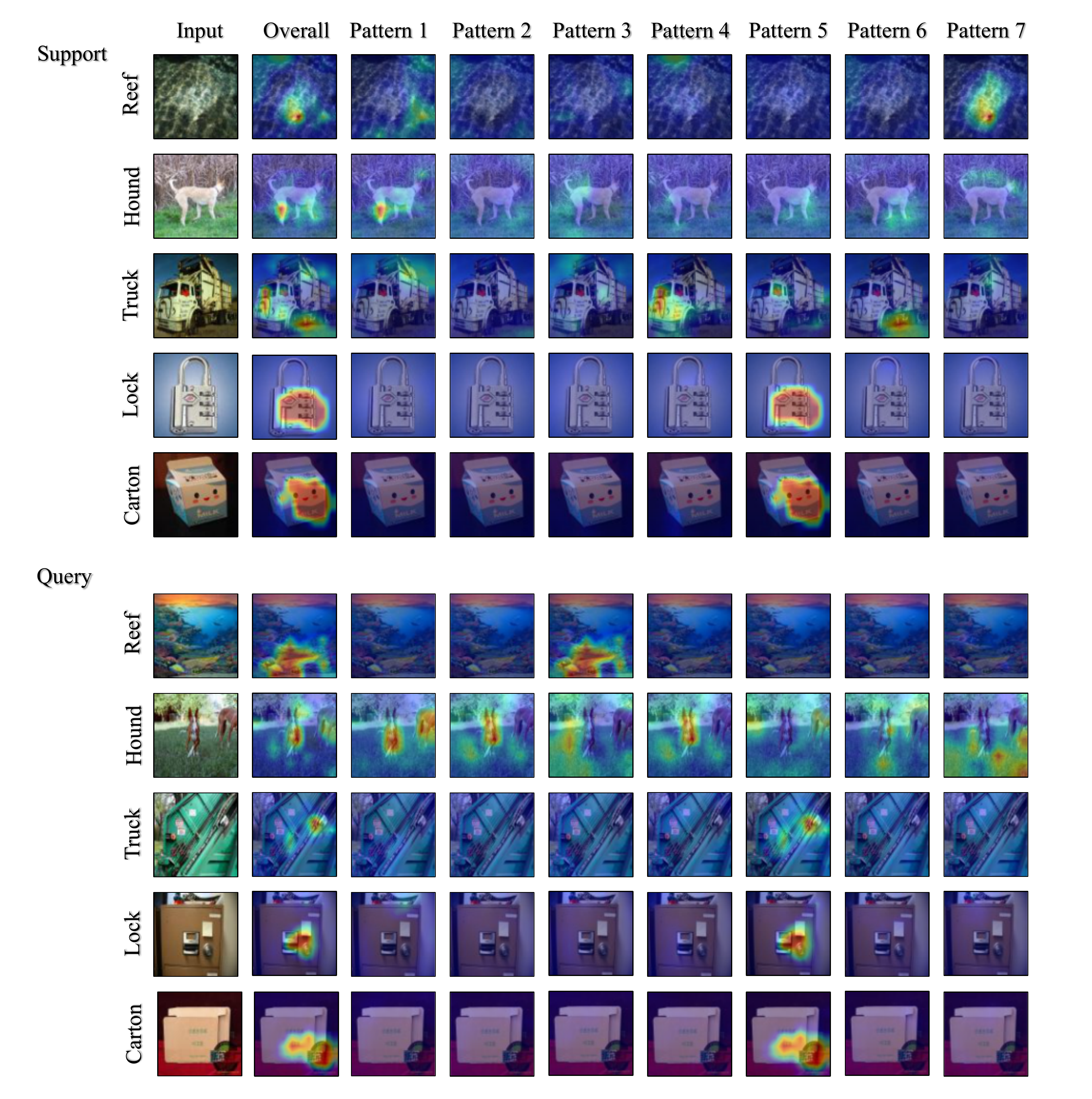}
	\caption{Pattern-based visualization of sample 4.}
	\label{pair4}
\end{figure*}

\begin{figure}[!t]
	\centering
	\includegraphics[width=1\columnwidth]{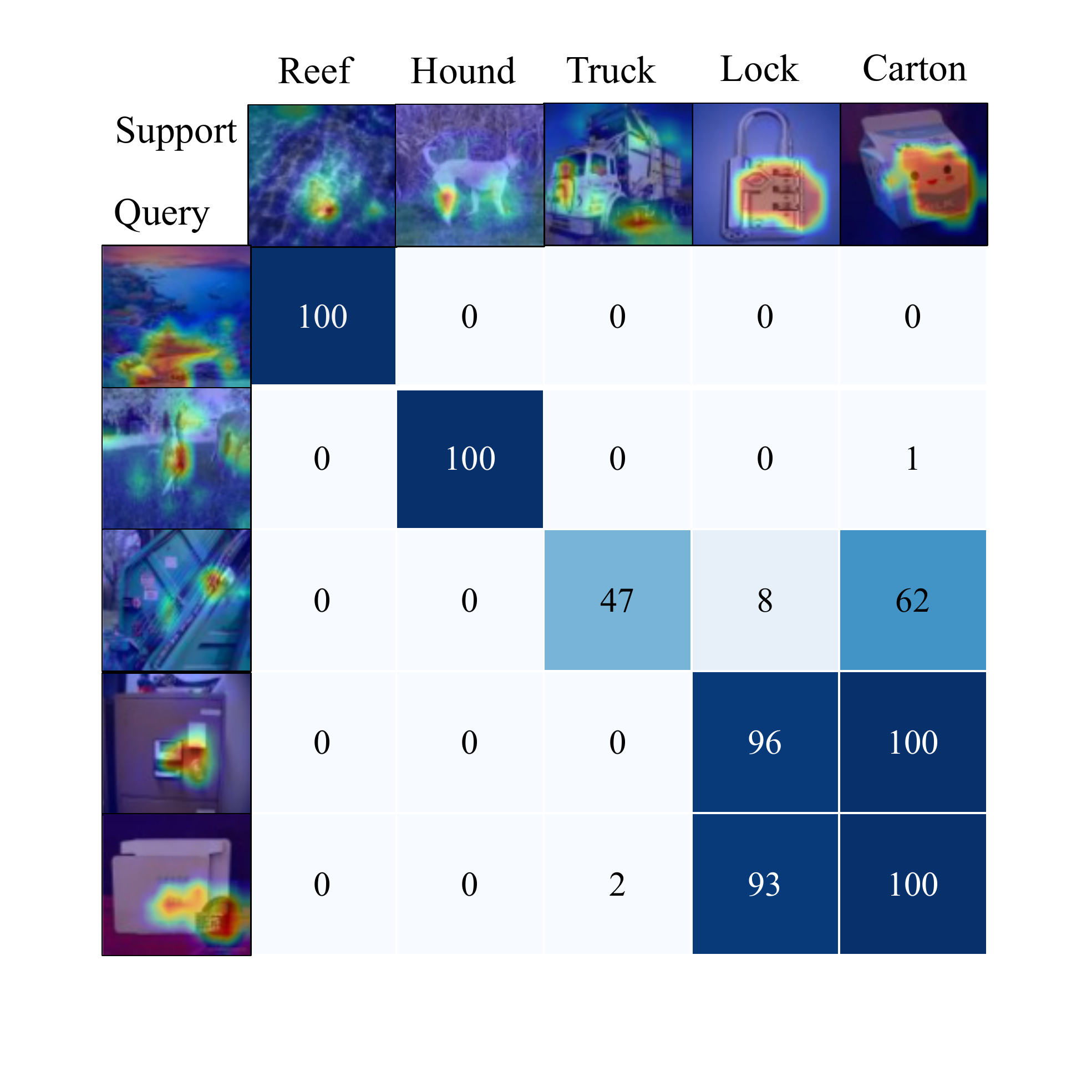}
	\caption{Pairwise matching of sample 1.}
	\label{matrix1}
\end{figure}

\begin{figure}[!t]
	\centering
	\includegraphics[width=1\columnwidth]{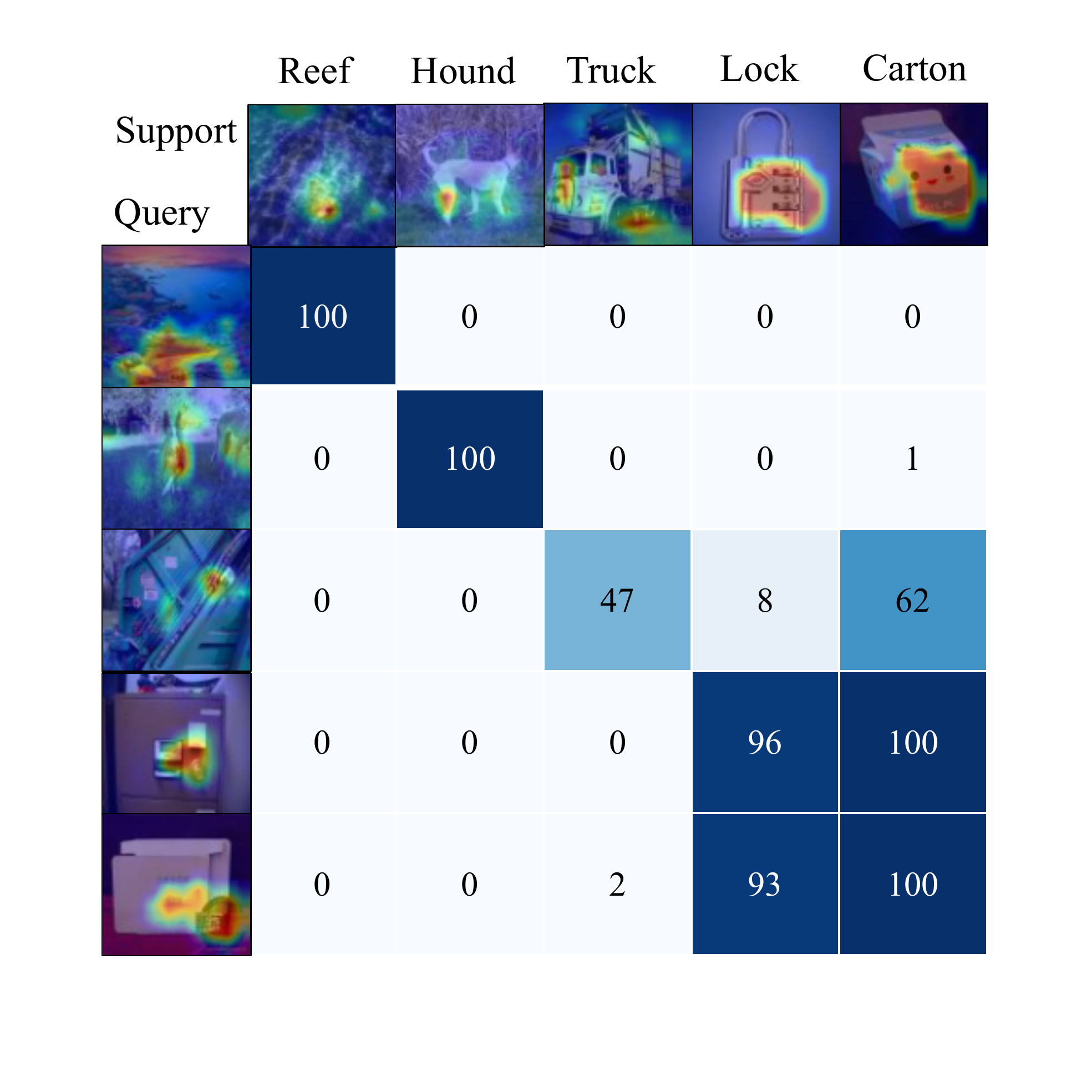}
	\caption{Pairwise matching of sample 2.}
	\label{matrix2}
\end{figure}

\begin{figure}[!t]
	\centering
	\includegraphics[width=1\columnwidth]{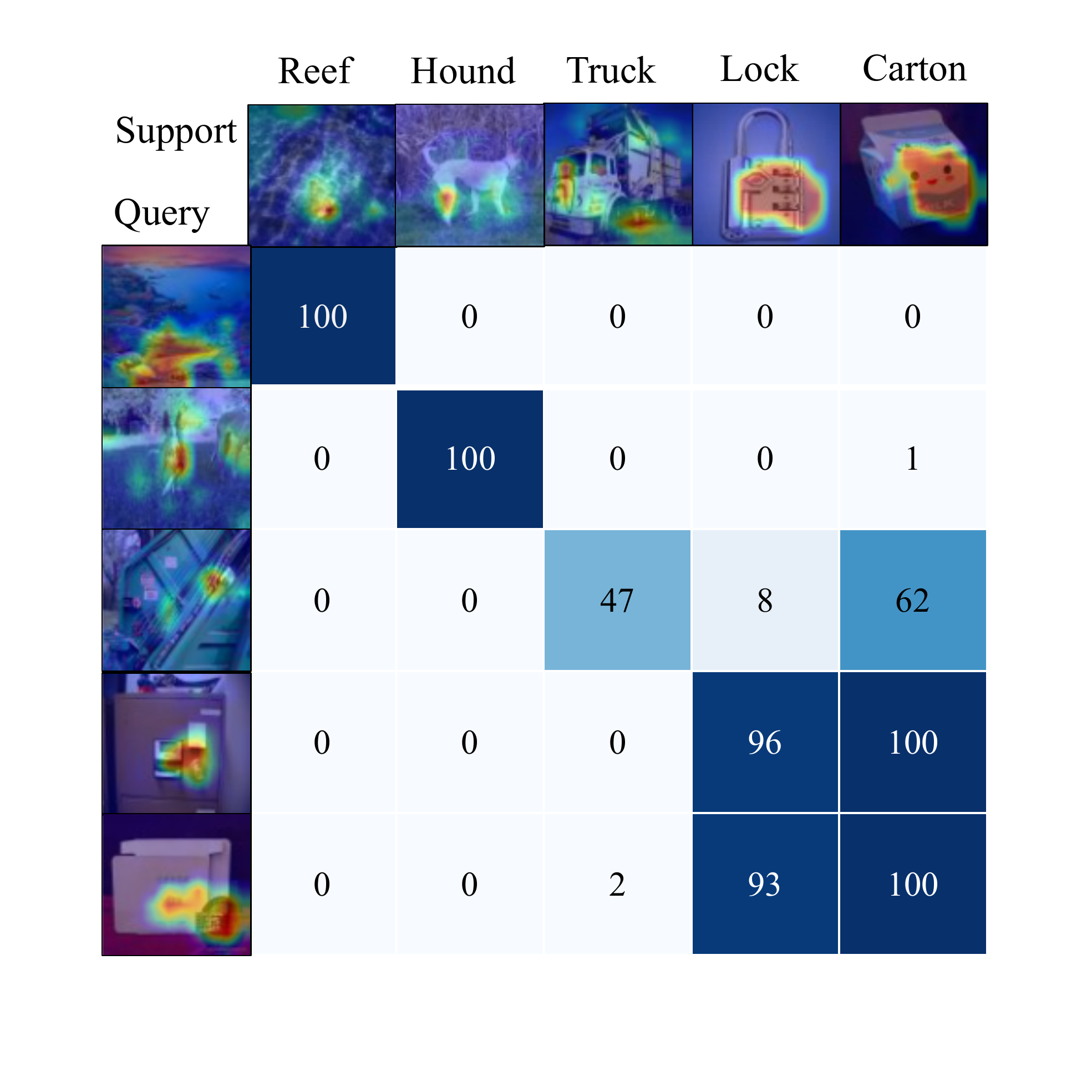}
	\caption{Pairwise matching of sample 3.}
	\label{matrix3}
\end{figure}

\begin{figure}[!t]
	\centering
	\includegraphics[width=1\columnwidth]{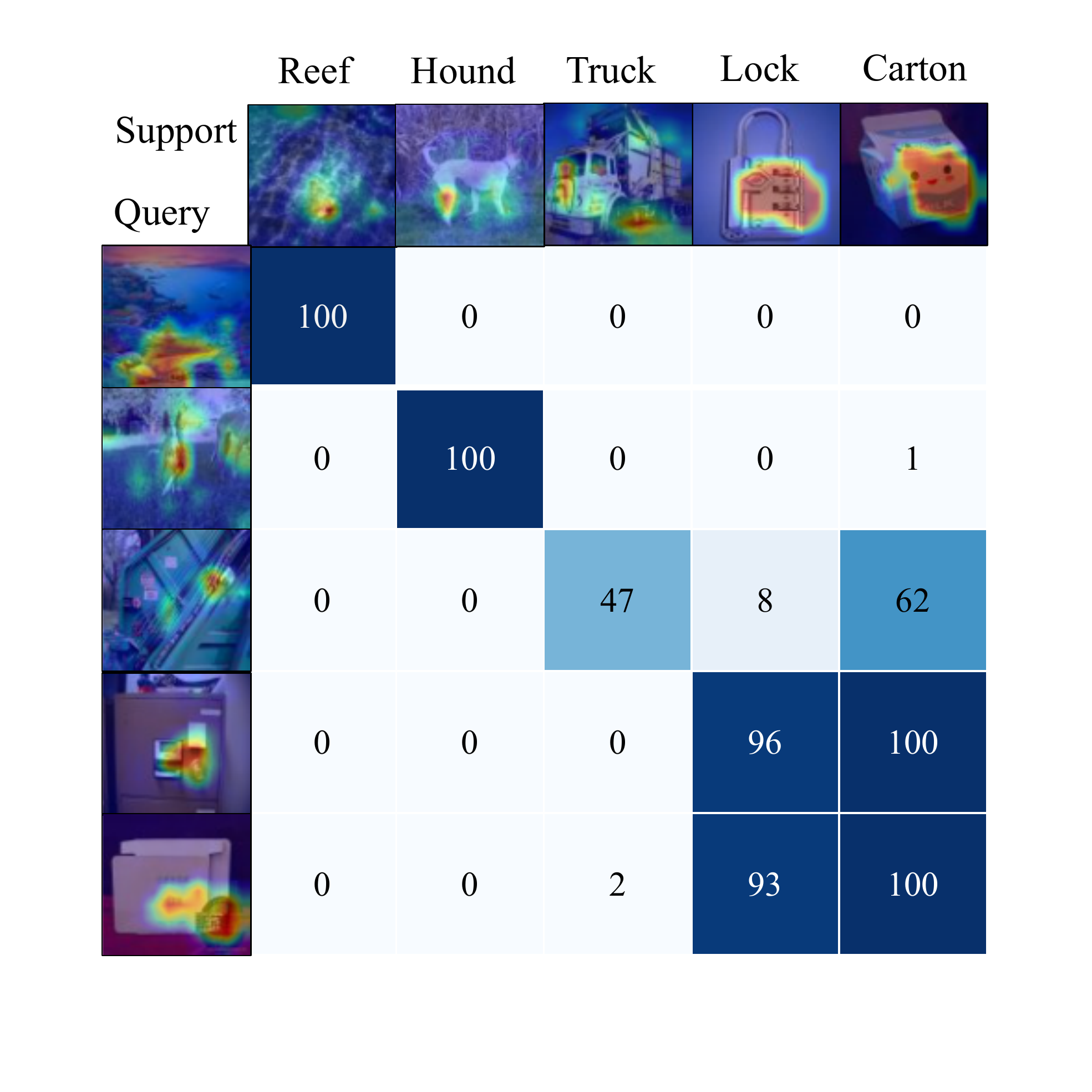}
	\caption{Pairwise matching of sample 4.}
	\label{matrix4}
\end{figure}

We provide visualization of patterns for 4 randomly sampled 5-way 1-shot tasks with a single query image over mini-ImageNet. The pattern-based visualization (Figures \ref{pair1}--\ref{pair4}) and the pairwise matching scores (Figures \ref{matrix1}--\ref{matrix4}) are shown for sample 1--4, respectively. We shall provide some discussion on the respective samples.

\paragraph{Sample 1}
By observing the matching matrix in Figure \ref{matrix1}, we find there are two confusing categories of \texttt{lock} and \texttt{carton}. They all get a high score for each other category. The visualization in Figure \ref{pair1} shows that pattern 5 is responsible for both the letters (or a face of the character) on the carton and the discs of the lock. We would say that the letters and the discs share some similar structures, which cause the confusion.

\paragraph{Sample 2}
The pairwise matching scores in this sample find proper matches except for \texttt{poncho}. In Figure \ref{matrix2}, the \texttt{poncho} image in support is a baby girl wearing a poncho, while the image in query is just the poncho with black color in the white background. The query image for \texttt{poncho} yields high scores for the support images of \texttt{poncho}, \texttt{skirt}, and \texttt{beetle}. The highest score of \texttt{beetle} may be due to the black color. Interestingly, the support and query images for \texttt{skirt} shows the attention over the door behind the person but not over the skirt itself. This is a good example of the importance of explanation for FSL.

\paragraph{Sample 3}
In Figure \ref{pair3}, we find both the query and support give attention on the body part of the goose, but the differences in the perspective and the number of objects may make matching difficult. As a result, the query \texttt{goose} gets low scores for all support images. This also happens to \texttt{carton} in this sample.

\paragraph{Sample 4}
In Figure \ref{matrix4}, the query \texttt{hound} shows high matching scores to hound, goose, catamaran, and skirt. Through Figure \ref{pair4}, we find that the query \texttt{hound} contains a hound as well as people, which is also in the supports \texttt{skirt} and \texttt{catamaran}. This means that some learned patterns cover people, and the people in the query \texttt{hound} lead to high matching scores against \texttt{skirt} and \texttt{catamaran}. Inclusion of different objects often causes prediction failure.

\typeout{get arXiv to do 4 passes: Label(s) may have changed. Rerun}
\end{document}